\documentclass[10.5pt, a4paper]{article}

\usepackage{geometry}
\usepackage{authblk}
\usepackage{apacite}
\usepackage{amsfonts}
\usepackage{amsthm}
\usepackage{amsmath}
\usepackage{graphicx}
\usepackage{subfigure}
\usepackage[colorlinks,linkcolor=red,citecolor=blue]{hyperref}
\usepackage{pifont}
\usepackage{natbib}
\usepackage{booktabs}
\usepackage{threeparttable}
\usepackage{algorithm}
\usepackage{algorithmic}
\usepackage{amssymb}
\usepackage{dsfont}
\usepackage{mathrsfs}
\usepackage{setspace}
\providecommand{\keywords}[1]{\textbf{\textit{Keywords:}} #1}

\title{A working likelihood approach to support vector regression with a data-driven insensitivity parameter}
\author[ ]{Jinran Wu}
\author[ ]{You-Gan Wang\thanks{Corresponding author. \\
  E-mail addresses: jinran.wu@hdr.qut.edu.au (J. Wu); you-gan.wang@qut.edu.au (Y-G. Wang).}}
\affil[ ]{School of Mathematical Sciences, Queensland University of Technology, Australia.}

\begin{document}

\maketitle
%\centerline{(Submitted to Machine Learning Journal)}

\setcounter{page}{1}

\raggedright
%\newpage

\begin{abstract}
The insensitive parameter in support vector regression determines the set of support vectors that greatly impacts the prediction. A data-driven approach is proposed to determine an approximate value for this insensitive parameter by minimizing a generalized loss function originating from the likelihood principle. This data-driven support vector regression also statistically standardizes samples using the scale of noises. Nonlinear and linear numerical simulations with three types of noises ($\epsilon$-Laplacian distribution, normal distribution, and uniform distribution), and in addition, five real benchmark data sets, are used to test the capacity of the proposed method. Based on all of the simulations and the five case studies, the proposed support vector regression using a working likelihood, data-driven insensitive parameter is superior and has lower computational costs.
\end{abstract}


%\newpage

\parindent 3mm

\section{Introduction}

In the machine learning field, support vector regression (SVR) has been popular in management and engineering applications \citep{trafalis2000support, mohandes2004support, vrablecova2018smart, li2017new}, due to its solid theoretical foundation \citep{vapnik1997support11, chang2011libsvm, chang2002training} and insensitivity to the dimensionality of the samples \citep{drucker1997support}. As recommended by \citet{vapnik2013nature}, the parameter settings in SVR modeling contribute the generalization of the predictive performance. However, practitioners applying SVR in real-world applications often cannot obtain the most effective model. There are two key approaches to setting the hyper-parameter. One option is to use the $\it{k}$-cross validation to choose the parameters for SVR \citep{hastie2005elements, ito2003optimizing}. The other approach is to set the parameter as a constant, based on the empirical practice developed by \citet{chang2011libsvm}. In particular, the researchers suggested that the regularization parameter $C$ and the insensitive parameter $\epsilon$ be set at $1.0$ and $0.1$, respectively. However, although the tuning parameter setting provides an acceptable generalization in most conditions, there is still a huge gap between this solution and the best SVR using the optimal parameters.

For the insensitive parameter $\epsilon$ that controls the number of support vectors \citet{scholkopf1999shrinking}, \citet{scholkopf2000new} used the parameter $\nu$ to effectively control the number of support vectors to eliminate the free parameter, $\epsilon$. However, one drawback is that the choice of $\nu$ has an impact on the generalization of the model \citep{scholkopf1998support}. Furthermore, insensitive parameter estimation methods that consider the noises in observations have been developed. \citet{jeng2003support} proposed to estimate the insensitive parameter in two steps. The first step is to estimate the regression errors by the SVR at $\epsilon=0$. Then, the $\epsilon$ value is updated by  $c \hat{\sigma}$ with an empirical constant $c$ and the estimated standard deviation of the noise $\hat{\sigma}$. In the absence of outliers, the standard deviation can be calculated based on all the regression errors, and $c$ is set as $1.98$. Otherwise, a trimmed estimator is obtained by removing $5-10\%$ of samples at both ends to achieve robustness, and $c$ is recommended to be fixed at 3. Obviously, although \citet{jeng2003support}'s method aims to incorporate data size in the estimation, the empirical settings make the method unable to recognize the noise level to estimate the insensitive parameter $\epsilon$. Similar to \citet{jeng2003support}'s method, \citet{cherkassky2004practical} incorporated sample size into the insensitive parameter estimation. As explored by them, the empirical formulation for $\hat{\epsilon}$ is calculated by the product of the empirical constant $3$,  the standard deviation of the noise, and an empirical coefficient $\sqrt{\ln n / n}$ ($n$ is the sample size). However, when the sample size increases, this $\hat{\epsilon}$ would approach to $0$, so this method does not recognize the noise level for the insensitive parameter estimation.

As explained by \citet{vapnik2013nature}, the insensitive loss function consists of the least modulus (LM) loss and the special Huber loss function when $\epsilon=0$. Hence, in our study, considering the insensitive Laplacian distribution loss function inspired by \citet{vapnik1997support11} and \citet{bartlett2002model}, we focused on the insensitive parameter $\epsilon$ and propose a novel SVR with a data-driven (D-D) insensitive parameter. Similar to  \citet{jeng2003support} and \citet{cherkassky2004practical}'s work, our method is developed on the theoretical background of SVR instead of parameter estimation based on re-sampling. Motivated by \citet{wang2007robust}, we propose designating the working likelihood to estimate the insensitive parameter for SVR. In other words, the working likelihood method can estimate appropriate hyper-parameters to find the most appropriate $\epsilon$-Laplacian distribution to the real noise distribution. Our working likelihood (or D-D) method works as a vehicle for the $\epsilon$ loss function parameter estimation. In addition, different from the computational standardization, the target in the proposed model is standardized in a statistical manner using the scale of the noise. Thus, our D-D method is more practicable and intelligent. In our simulations (linear and nonlinear), three types of error distributions were used to test the D-D insensitive parameter estimation, namely, the insensitive Laplacian distribution, normal distribution, and uniform distribution. Furthermore, some case studies were applied to validate that our D-D SVR has novel generalization in real applications.

This rest of this paper is organized as follows. Section \ref{ddsvr} reviews the framework of SVR and outlines the working likelihood for insensitive parameter estimation in SVR. Numerical simulations for three different types of noise sources (the insensitive Laplacian distribution, normal distribution, and uniform distribution) were implemented, and Section \ref{simulation} presents a discussion of the analyses of the simulation results, which proved the efficiency of the working likelihood. Then, in Section \ref{case}, we discuss the validation of our D-D SVR on five real data sets: energy efficiency, Boston housing, yacht hydrodynamics, airfoil self-noise, and concrete compressive strength. Finally, in Section \ref{conclusion}, we summarize the results that indicate the working likelihood (D-D) method has superior performance on insensitive parameter estimation based on the real noise information in SVR, indicating that our D-D SVR is very promising for predictions.

\section{Data-driven support vector regression (SVR)}\label{ddsvr}

\subsection{The framework of SVR}

Assume the training data ${(x_1, \ y_1),...,(x_n, \ y_n)}\in \mathsf{\chi}\times \mathbb{R}$, where $\mathsf{\chi}$ denotes the space of the input patterns. In $\epsilon$-SVR, the target is to obtain a function $f(x)$ that has at most $\epsilon$ deviation from the actual obtained target $y_i$ for all the training data, and at the same time, is as flat as possible \citep{smola2004tutorial,drucker1997support}. This means that smaller errors ($\leq \epsilon$) are ignored, and larger errors will be accounted for in the loss function. The case of linear function $f(\cdot)$ can be formed as
\begin{equation}
f(x)=\langle \omega, x \rangle+b \quad \omega \in \mathsf{\chi}, b \in \mathbb{R},
\label{regression_model}
\end{equation}
where $\langle .,. \rangle$ represents the dot product in $\mathsf{\chi}$. Flatness in Eq. (\ref{regression_model}) means finding a small $\epsilon$. Now we are interested in minimizing the Euclidean norm, meaning ${\Vert \omega \Vert}^{2}$, which can be expressed with a convex optimization problem as \citep{smola2004tutorial},
\begin{equation}
\begin{aligned}
\mbox{minimize} \ \ & \frac{1}{2}\Vert \omega \Vert ^2 \\
\mbox{subject to} \ \ & \left \{
           \begin{array}{l}
           y_i - \langle \omega, x_i\rangle-b \leqslant \epsilon,\\
           \langle \omega, x_i\rangle + b -y_i \leqslant \epsilon.\\
           \end{array} \right.
\end{aligned}
\label{original_optimization}
\end{equation}
Here, the optimization problem is feasible; it means that there exists such a function $f$ that approximates all pairs ($x_i, \ y_i$) with $\epsilon$ precision. Then, the slack variables $\xi_i$ and $\xi_i^*$ are introduced to cope with the otherwise infeasible constraints of the optimization version in Eq. (\ref{original_optimization}). Hence, the formulation is shown as,
\begin{equation}
\begin{aligned}
\mbox{minimize} \ \ & \frac{1}{2}\Vert \omega \Vert ^2 + C \sum \limits_{i=1}^{n} (\xi_i+\xi_i^*)\\
\mbox{subject to} \ \ & \left \{
           \begin{array}{l}
           y_i - \langle \omega, x_i\rangle-b \leqslant \epsilon+\xi_i,\\
           \langle \omega, x_i\rangle + b -y_i \leqslant \epsilon+\xi_i^*,\\
           \xi_i,\xi_i^*\geqslant 0.
           \end{array} \right.
\end{aligned}
\label{constraint_optimization}
\end{equation}
The regularization parameter $C$ (a positive constant) determines the trade-off between the flatness of $f$ and the amount up to which deviations are larger than $\epsilon$. The optimization problem can be transformed to its dual problem as follows \citep{smola2004tutorial}:
\begin{equation}
\begin{aligned}
\mbox{maximize} \ \ & - \frac{1}{2} \sum \limits _{i,j=1}^{n} (\alpha_i-\alpha_i^*)(\alpha_j-\alpha_j^*)\langle             x_i, x_j \rangle \\
 \quad \quad   &-\epsilon \sum \limits_{i=1}^{n}(\alpha_i+\alpha_i^*)+\sum \limits_{i=1}^{n} y_i (\alpha_i-\alpha_i^*)\\
\mbox{subject to} \ \ & \left \{
           \begin{array}{l}
           \sum \limits_{i=1}^n (\alpha_i-\alpha_i^*)=0,\\
            \alpha_i,\alpha_i^* \in [0,C].\\
           \end{array} \right.
\end{aligned}
\label{dual_optimization}
\end{equation}
Here, $\alpha_i$ and $\alpha_i^*$ are Lagrange multipliers for $\epsilon+\xi_i-y_i+\langle \omega, x_i \rangle+b$ and $\epsilon+\xi_i^*-\langle \omega, x_i \rangle-b+y_i$, respectively. This dual optimization has a general solution,
\begin{equation}
f(x)=\sum \limits_{i=1}^{n} (\alpha_i-\alpha_i^*)k(x_i,x)+b,
\end{equation}
where the dual optimization is subjected to the constraints $0 \leqslant \alpha_i,\alpha_i^* \leqslant C$, and  $k(x_i, x)$ is the  kernel function including linear function  as a special case.

As illustrated by \citet{vapnik2013nature}, three important parameter settings in SVR significantly impact the model's generalization: the regularization parameter $C$, the kernel parameters, and the insensitive parameter $\epsilon$. The first one, $C$, can be estimated by
the  0.95 quantile of $\vert y_i \vert$ \citep{cherkassky2004practical},
\begin{equation}
C_{CM}=\vert y_i \vert _{(0.95)},i=1,...,n.
\end{equation}

Then, the second kernel parameter is applied to adjust the mapping from the original space to the high-dimensional space; this is decided by the type of kernel function and the application domain. The last one is the most important parameter, $\epsilon$, which controls the number of support vectors. We will explore how to estimate the insensitive parameter $\epsilon$ based on the loss function mechanism from a statistical perspective in the next section.

\subsection{Working likelihood for insensitive parameter estimation}\label{data-driven_estimation}

Suppose the training data set consists of $n$ samples $(x_i, y_i), (i=1, 2, ..., n)$, and the target $y_i$, is generated from the following model:
\begin{equation}
   y_i = f(x_i) + s  u_i,
\end{equation}
where $f(\cdot)$ represents the expected value, while the second component, $s  u_{i}$ (which is denoted by $U_i$) is the noise ($s$ is the scale, and $u_i$ is the noise after scaling $s$).

In $\epsilon$-SVR, the loss function is defined as
\begin{equation}
\begin{aligned}
V(u) & =\vert u \vert _{\epsilon},\\
               & =\left\{
               \begin{array}{lrl}
               u -\epsilon & & {u >\epsilon},\\
               0                     & & {-\epsilon \leqslant u \leqslant \epsilon},\\
               -u -\epsilon & & {u < -\epsilon}, \\
               \end{array} \right.
\end{aligned}
\label{loss_defin}
\end{equation}
where $u=y-\langle \omega, x \rangle -b$ is the residual item. The corresponding density function for $u_i$ is,
\begin{equation}
g(u; \epsilon)=\frac{1}{2(1+\epsilon)} \exp(-\vert u \vert_{\epsilon}),
\end{equation}
which will correspond to the loss function given by Eq. (\ref{loss_defin}) up to a constant.

Thus, suppose that all $u_i$ are identically and independently distributed with a density function $g(\cdot)$. Let $\theta$ be a vector collecting all the unknown parameters $(\epsilon, s)$.
  The negative log-likelihood based on the training data is then

\begin{equation}
-\log   L (\theta)=  - \sum_{i=1}^n \log \left(  g \left( \frac{y_i- f(x_i)} {s} \right)   \right) +  n \log(s).
\end{equation}

Once the SVR approach is adapted, we essentially assume $u_i$ follows a density function that is proportional to $\exp(-V(u))$. Our working likelihood D-D method estimates all the parameters in $\theta$ by maximizing $L(\theta)$.
In particular, we investigate the choice of the insensitivity parameter $\epsilon$ in the SVR approach. Clearly, the $\epsilon$ value that results by maximizing $L$ is data dependent and expected to be more effective. Meanwhile, the scale of the noise $s$ can also be estimated.

Next, recalling that $U_i=s u_i$, assume that $U_{1}, U_{2}, \ldots U_{n}$ are independent and identically distributed random variables. Denote $\left( \epsilon, s \right)=\theta$. Their joint working likelihood function is
\begin{equation}
L ( \theta )= \prod \limits _ {i=1}^{n} \left(  \frac{1}{s}  g(\frac{U_i}{s}; \epsilon, s) \right) =\left(\frac{1}{s}\right)^n \cdot \left(\frac{1}{2(1+\epsilon)}\right) ^{n} \cdot \exp \left(  -\sum \limits _{i=1}^{n} \vert \frac{U_i}{s} \vert_{\epsilon} \right).
\end{equation}
Therefore, $L (\theta )$ is a likelihood function with  parameters $\epsilon$ and $s$ properly regularized. Their estimators can thus
be achieved by minimizing the negative log-likelihood function,
\begin{equation}
\label{LLF}
\begin{aligned}
\displaystyle -\log L(\theta) &=n \log s +n \log \left[ 2(1+\epsilon) \right]+\sum \limits _{i=1}^{n} \vert \frac{U_i}{s}\vert _{\epsilon} \\
\displaystyle&=n \log s +n \log \left(  2(1+\epsilon) \right) +\sum \limits _{i=1}^{n} \left( (\frac{U_i}{s}-\epsilon) \cdot \mathbb{I} (\frac{U_i}{s} >\epsilon)+ (-\frac{U_i}{s}-\epsilon) \cdot \mathbb{I} (\frac{U_i}{s} <-\epsilon) \right).
\end{aligned}
\end{equation}

The derivatives of $\left(-\log L(\theta) \right)$ with respect to $\epsilon$ and $s$ are given as
\begin{equation}
\begin{aligned}
\left\{
\begin{array}{l}
\displaystyle \frac{\partial \left( -\log L(\theta) \right)}{\partial \epsilon}=      \frac{n}{1+\epsilon}-\sum \limits _{i=1}^{n} \mathbb{I} ( \vert \frac{U_i}{s} \vert >\epsilon),\\
\displaystyle\frac{\partial \left( -\log L(\theta) \right)}{\partial s}= \frac{n}{s}-\frac{1}{s^2}\sum \limits_{i=1}^{n} \vert U_i \vert \cdot \mathbb{I} ( \vert \frac{U_i}{s} \vert >\epsilon).
\end{array} \right.
\end{aligned}
\label{rootd}
\end{equation}
By equating them to 0, both parameters $(\epsilon, s)$ can be expressed as,
\begin{equation}
\begin{aligned}
\left\{
\begin{array}{l}
{\epsilon} =
 \frac{\displaystyle \sum \limits_{i=1}^n \mathbb{I} (\vert \frac{ U_i}{s} \vert \leqslant \epsilon)}
     {\displaystyle \sum \limits_{i=1}^n \mathbb{I} (\vert \frac{ U_i}{s} \vert > \epsilon)},\\
s = \frac{\displaystyle \sum \limits _{i=1}^n \vert   U_i \vert \cdot \mathbb{I}(\vert  \frac{ U_i}{s} \vert > \epsilon)}{\displaystyle n}.
\end{array} \right.
\end{aligned}
\label{parameter_estimation}
\end{equation}
Thus, the parameters $\epsilon$ and $s$ can be estimated by minimizing Eq. (\ref{LLF}) or calculating the root of Eq. (\ref{parameter_estimation}). In addition, the meaning of $(\epsilon, s)$ now becomes clear. This indicates that $\epsilon$ is the odds ratio of  being inside the box ($\leqslant \epsilon$) versus outside the box ($\geqslant \epsilon$). The parameter $s$ is the average distance of the support vectors, while the distance of non-support vectors is regarded as 0.

\begin{figure}[htpb]
\centering
\includegraphics[height=21cm,width=14cm]{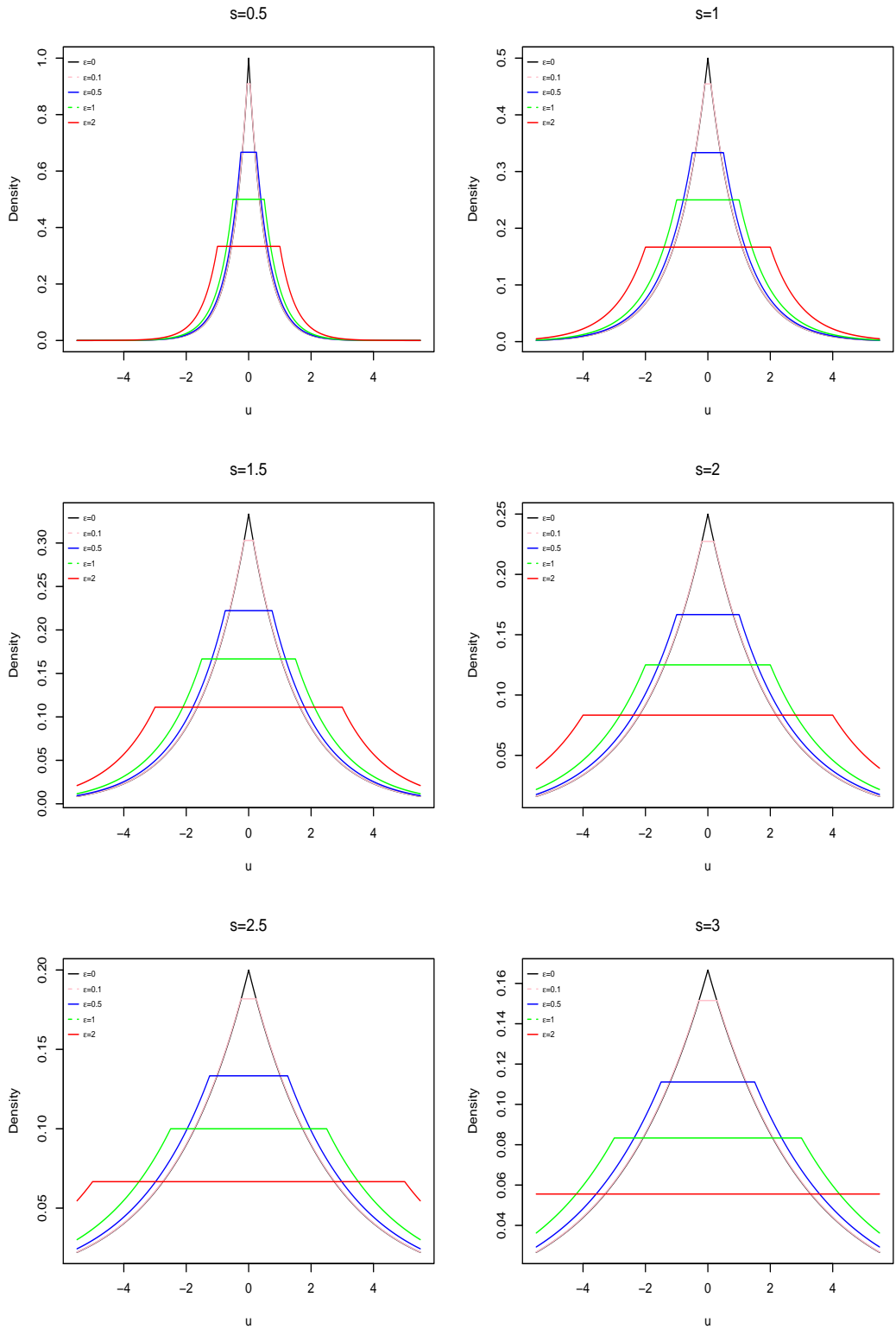}
\caption{Working likelihood functions with different insensitive parameters at different scales.}
\label{workinglikelihood_key}
\end{figure}

As $n\rightarrow \infty$, we can obtain the limiting values of $\epsilon$ and $s$ for a given distribution of noise $u_i$. Suppose that $g(\cdot)$ is the density function of the noise term $u_i$. Asymptotically, Eq. (\ref{parameter_estimation}) becomes,

\begin{equation}
\begin{aligned}
\left\{
\begin{array}{l}
\displaystyle \frac{1}{\epsilon^*+1}=Pr ( \vert \frac{U}{s^*} \vert > \epsilon^{*}),\\
\displaystyle 1= \int_{\epsilon^{*}}^{+\infty} u \left( g(u)-g(-u) \right)du.
\end{array} \right.
\end{aligned}
\label{limited_estimation}
\end{equation}
Each paired $\theta=\left(\epsilon, s \right)$ value corresponds to a potential key to a real data set. We now propose obtaining the``best" key in the tool box. Figure \ref{workinglikelihood_key} shows some potential keys for inferring the unknown noise. This means the $\epsilon$-Laplacian distribution can approximate the real noise distribution by adapting the scale parameter $s$ and the insensitive parameter $\epsilon$.

Finally, the framework of our D-D SVR with D-D insensitive parameters can be given as follows:

\begin{enumerate}

\item[Step 1.] Apply the $\epsilon$-SVR ($\it{\epsilon} =0$, $\it{C} = 1$) in training sets, and obtain residuals $U_i$;

\item[Step 2.] Estimate the insensitive parameter $\epsilon$ and the scale parameter $s$ by minimizing Eq. (\ref{LLF});

\item[Step 3.] Train our D-D SVR using the updated $\hat{\epsilon}$ and $\hat{s}$; and

\item[Step 4.] Predict the targets in the test set.

\end{enumerate}

\section{Simulation experiments}\label{simulation}

To illustrate how the working likelihood produces D-D parameter estimation (D-D) and a prediction, we now consider three types of residuals generated from the uniform distribution, the norm distribution, and the $\epsilon$-Laplacian distribution, respectively.

For comparison, we will investigate other three insensitive parameter estimation methods for the $\epsilon$-SVR. The first one is the tuning parameter setting (tuning) ($C=1.0$ and $\epsilon=0.1$)  \citep{chang2011libsvm}. The second method, \citet{cherkassky2004practical}'s empirical parameter approach (CM), is
\begin{equation}
\epsilon_{CM} = 3 \sigma_{\mbox{noise}} \sqrt{\frac{\ln n}{n}},
\end{equation}
where the standard deviation of noise $\sigma_{\mbox{noise}}$ is obtained from the residuals using $\epsilon=0$. The last one is the $\it{k}$-cross validation ($\it{k}$-CV), where $\it{k}$ is fixed at $10$, and $5$ alternative $\epsilon$ settings are set as $0.01, 0.05, 0.1, 0.2$ and $0.3$. Both mean absolute error (MAE) and root mean square error (RMSE) are calculated for comparison.
\begin{equation}
\mbox{MAE}=\frac{1}{n}\sum \limits_{i=1}^{n} \vert \hat{y}_{i}-y_i \vert,
\end{equation}
\begin{equation}
\mbox{RMSE}=\sqrt{\frac{1}{n}\sum \limits_{i=1}^{n} (\hat{y}_{i}-y_i)^2},
\end{equation}
where $\hat{y}_{i}$ is the $i$-th prediction, and $y_{i}$ is the $i$-th observation. For each method $\it{X}$ using the tuning method as the benchmark approach, two ratios are defined as
\begin{equation}
\mbox{ratio}_{\mbox{RMSE}}=\frac{\mbox{RMSE}_{\mbox{tuning}}}{\mbox{RMSE}_{\it{X}}},
\end{equation}
\begin{equation}
\mbox{ratio}_{\mbox{MAE}}=\frac{\mbox{MAE}_{\mbox{tuning}}}{\mbox{MAE}_{\it{X}}}.
\end{equation}
It is obvious that the method $\it{X}$ beats the tuning setting only if the ratio is larger than 1, and otherwise, it does not. The nonlinear simulations and linear simulations are applied to show the efficiency of our proposed D-D SVR.

\subsection{Nonlinear regression}

To demonstrate the performance of our D-D SVR for nonlinear system modeling, the univariate $sinc$ target function from the SVR literature \citep{drucker1997support, chu2004bayesian, xu2012weighted, karal2017maximum} is considered as
\begin{equation}
\begin{aligned}
y_i=a \cdot \frac{\sin (x_i)}{x_i}+s u_i, \quad i=1, 2, \ldots, n,
\end{aligned}
\end{equation}
where $x_i$ is generated from the uniform distribution $unif[-10, 10]$; $s$ is the scale of the noise level; and the standard noise $u_i$ is generated from a known distribution ($\epsilon$-Laplacian distribution, normal distribution $N(0, \sigma^2)$, and uniform distribution $unif[-b, b]$). In addition, to make our simulations more meaningful, the scale of nonlinear system $a$ is set as $5$, $4$, and $6$ from insensitive-Laplacian noises, normal noises, and uniform noises, respectively. Also, we generate $\it{n}$ simulation samples, and then the samples are divided into two groups of the same size. All experiments are repeated $100$ times to calculate the average performance of the benchmark SVRs and our proposed D-D SVR. The kernel of the SVR is the default radial basic function. It should be noted that, for our comparison, the ratio is calculated based on the gap between the prediction $\hat{y}_i$ and the $\mu_i$ ($\mu_{i}=a\sin (x_i)/ x_i$). This can show the performance of our D-D SVR at eliminating the interruption from noise and model a real system. All of the nonlinear simulation results are displayed in Table \ref{nonlinearLaplacian} (insensitive Laplacian distribution), Table \ref{nonlinearnorm} (normal distribution), and Table \ref{nonlinearuniform} (uniform distribution).

\begin{table}[htbp]
\footnotesize
\centering
\caption{Nonlinear case ($\epsilon$-Laplacian distribution): Relative performance of the CM, $10$-CV, and D-D methods in comparison to the tuning approach.}
\label{nonlinearLaplacian}
\begin{tabular}{rrcccccccccc}
\toprule
\multicolumn{3}{l}{Noise settings}&\multicolumn{2}{c}{Parameters}&\multicolumn{2}{c}{$\mbox{CM}$}&\multicolumn{2}{c}{$\mbox{$10$-CV}$}&\multicolumn{2}{c}{$\mbox{D-D}$}\\
  \cmidrule(lrr){1-3} \cmidrule(lrr){4-5}	\cmidrule(lr){6-7} \cmidrule(lr){8-9} \cmidrule(lr){10-11}
$\it{n}$&$s$& $\epsilon$& $\hat{s}$&$\hat{\epsilon}$& $\mbox{ratio}_{\mbox{MAE}}$& $\mbox{ratio}_{\mbox{RMSE}}$& $\mbox{ratio}_{\mbox{MAE}}$& $\mbox{ratio}_{\mbox{RMSE}}$& $\mbox{ratio}_{\mbox{MAE}}$& $\mbox{ratio}_{\mbox{RMSE}}$\\
\midrule
200&	0.8&	0.2&	0.66&	0.00&	0.95&	0.97&	0.99&	1.00&	1.47&	1.32\\
400&	0.8&	0.2&	0.73&	0.02&	0.94&	0.95&	1.11&	1.00&	1.71&	1.55\\
1000&	0.8&	0.2&	0.77&	0.01&	0.95&	0.95&	1.11&	1.01&	1.86&	1.69\\
200&	0.8&	0.5&	0.70&	0.00&	0.96&	0.97&	1.07&	1.01&	1.44&	1.36\\
400&	0.8&	0.5&	0.77&	0.04&	0.95&	0.96&	1.09&	1.00&	1.66&	1.52\\
1000&	0.8&	0.5&	0.82&	0.06&	0.96&	0.96&	1.10&	1.01&	1.70&	1.56\\
200&	0.8&	1.0&	0.81&	0.03&	0.96&	0.97&	1.06&	1.00&	1.30&	1.22\\
400&	0.8&	1.0&	0.87&	0.15&	0.97&	0.97&	1.09&	1.00&	1.54&	1.42\\
1000&	0.8&	1.0&	0.87&	0.53&	0.99&	0.99&	1.10&	1.01&	1.71&	1.56\\
\hline
200&	1.0&	0.2&	0.85&	0.00&	0.96&	0.97&	1.10&	1.01&	1.47&	1.34\\
400&	1.0&	0.2&	0.93&	0.01&	0.94&	0.95&	1.10&	1.00&	1.66&	1.51\\
1000&	1.0&	0.2&	0.98&	0.01&	0.94&	0.94&	1.10&	1.01&	1.78&	1.64\\
200&	1.0&	0.5&	0.89&	0.00&	0.95&	0.96&	1.07&	0.99&	1.41&	1.31\\
400&	1.0&	0.5&	0.97&	0.02&	0.95&	0.96&	1.08&	1.00&	1.55&	1.44\\
1000&	1.0&	0.5&	1.03&	0.05&	0.96&	0.96&	1.08&	1.01&	1.65&	1.54\\
200&	1.0&	1.0&	0.99&	0.04&	0.98&	0.98&	1.05&	1.00&	1.24&	1.18\\
400&	1.0&	1.0&	1.08&	0.15&	0.97&	0.98&	1.05&	1.00&	1.42&	1.35\\
1000&	1.0&	1.0&	1.08&	0.57&	1.00&	1.00&	1.09&	1.01&	1.67&	1.54\\
\hline
200&	1.2&	0.2&	1.02&	0.00&	0.94&	0.95&	1.08&	1.00&	1.39&	1.28\\
400&	1.2&	0.2&	1.10&	0.00&	0.93&	0.94&	1.09&	1.00&	1.55&	1.41\\
1000&	1.2&	0.2&	1.17&	0.01&	0.93&	0.93&	1.11&	1.01&	1.73&	1.58\\
200&	1.2&	0.5&	1.08&	0.02&	0.95&	0.95&	1.06&	1.00&	1.26&	1.19\\
400&	1.2&	0.5&	1.16&	0.01&	0.95&	0.95&	1.07&	0.99&	1.45&	1.34\\
1000&	1.2&	0.5&	1.24&	0.06&	0.96&	0.97&	1.08&	1.01&	1.64&	1.52\\
200&	1.2&	1.0&	1.20&	0.07&	0.99&	0.98&	1.04&	1.00&	1.18&	1.14\\
400&	1.2&	1.0&	1.32&	0.14&	0.99&	0.99&	1.06&	1.01&	1.29&	1.22\\
1000&	1.2&	1.0&	1.31&	0.49&	1.00&	1.01&	1.09&	1.02&	1.52&	1.42\\
\bottomrule
\end{tabular}
\end{table}

As illustrated in Table \ref{nonlinearLaplacian}, compared with the CM and $10$-CV, the ratios of the D-D from both RMSE and MAE are significantly greater than $1$, indicating that our proposed SVR allowed for remarkable improvements in the forecasting performance for all 27 simulations. However, the insensitive parameter $\epsilon$ tends to be underestimated. The main reason for this is that, as shown in Figure \ref{workinglikelihood_key}, the scale mainly contributes to the working likelihood function when the insensitive parameter is small. Another reason is that the training sample size is not large enough to estimate the insensitive parameter accurately. As the training set size enlarges, the estimated insensitive parameter converges to the true $\epsilon$.

\begin{table}[htbp]
\footnotesize
\centering
\caption{Nonlinear case (normal distribution): Relative performance of the CM, $10$-CV, and D-D methods in comparison to the tuning approach.}
\label{nonlinearnorm}
\begin{tabular}{rrccccccccc}
\toprule
\multicolumn{3}{l}{Noise settings}&\multicolumn{2}{c}{Parameters}&\multicolumn{2}{c}{$\mbox{CM}$}&\multicolumn{2}{c}{$\mbox{$10$-CV}$}&\multicolumn{2}{c}{$\mbox{D-D}$}\\
  \cmidrule(lrr){1-3} \cmidrule(lrr){4-5}	\cmidrule(lr){6-7} \cmidrule(lr){8-9}\cmidrule(lr){10-11}
$\it{n}$&$s$& $\sigma$& $\hat{s}$&$\hat{\epsilon}$  & $\mbox{ratio}_{\mbox{MAE}}$& $\mbox{ratio}_{\mbox{RMSE}}$& $\mbox{ratio}_{\mbox{MAE}}$& $\mbox{ratio}_{\mbox{RMSE}}$& $\mbox{ratio}_{\mbox{MAE}}$& $\mbox{ratio}_{\mbox{RMSE}}$\\
\midrule
200&	0.7&	0.5&	0.22&	0.00&	0.96&	0.98&	1.00&	1.00&	1.37&	1.16\\
400&	0.7&	0.5&	0.24&	0.00&	0.97&	0.97&	1.01&	1.01&	1.67&	1.46\\
1000&	0.7&	0.5&	0.24&	0.47&	1.00&	1.00&	1.03&	1.03&	1.64&	1.48\\
200&	0.7&	1.0&	0.44&	0.00&	0.97&	0.98&	1.00&	1.00&	1.29&	1.20\\
400&	0.7&	1.0&	0.49&	0.06&	0.97&	0.98&	1.00&	1.00&	1.48&	1.36\\
1000&	0.7&	1.0&	0.45&	0.88&	0.98&	0.99&	1.01&	1.01&	1.50&	1.38\\
200&	0.7&	1.5&	0.66&	0.03&	0.98&	0.98&	1.00&	1.00&	1.17&	1.12\\
400&	0.7&	1.5&	0.70&	0.31&	0.97&	0.98&	1.00&	1.00&	1.38&	1.29\\
1000&	0.7&	1.5&	0.65&	1.05&	0.98&	0.99&	1.01&	1.01&	1.48&	1.36\\
\hline
200&	0.9&	0.5&	0.28&	0.00&	0.97&	0.99&	1.00&	1.01&	1.42&	1.25\\
400&	0.9&	0.5&	0.31&	0.01&	0.96&	0.97&	1.00&	1.01&	1.58&	1.42\\
1000&	0.9&	0.5&	0.30&	0.68&	0.99&	0.99&	1.02&	1.02&	1.59&	1.45\\
200&	0.9&	1.0&	0.57&	0.00&	0.97&	0.97&	1.00&	1.00&	1.17&	1.10\\
400&	0.9&	1.0&	0.62&	0.17&	0.97&	0.98&	1.00&	1.01&	1.33&	1.23\\
1000&	0.9&	1.0&	0.57&	0.90&	0.98&	0.99&	1.00&	1.00&	1.48&	1.36\\
200&	0.9&	1.5&	0.86&	0.10&	0.98&	0.98&	1.00&	1.00&	1.09&	1.04\\
400&	0.9&	1.5&	0.89&	0.40&	0.98&	0.99&	1.00&	1.01&	1.25&	1.19\\
1000&	0.9&	1.5&	0.81&	1.15&	0.99&	0.99&	1.01&	1.01&	1.37&	1.28\\
\hline
200&	1.1&	0.5&	0.35&	0.00&	0.96&	0.97&	1.00&	1.00&	1.36&	1.24\\
400&	1.1&	0.5&	0.38&	0.04&	0.97&	0.97&	1.00&	1.00&	1.56&	1.41\\
1000&	1.1&	0.5&	0.35&	0.85&	0.98&	0.98&	1.01&	1.01&	1.59&	1.45\\
200&	1.1&	1.0&	0.69&	0.03&	0.98&	0.98&	1.01&	1.01&	1.14&	1.08\\
400&	1.1&	1.0&	0.75&	0.29&	0.98&	0.98&	1.00&	1.00&	1.33&	1.24\\
1000&	1.1&	1.0&	0.68&	1.03&	0.99&	0.99&	1.00&	1.00&	1.42&	1.31\\
200&	1.1&	1.5&	1.03&	0.17&	1.00&	1.00&	1.00&	1.00&	1.01&	0.98\\
400&	1.1&	1.5&	1.06&	0.56&	0.98&	0.98&	1.00&	1.00&	1.17&	1.11\\
1000&	1.1&	1.5&	1.01&	1.07&	0.99&	0.99&	1.00&	1.00&	1.30&	1.22\\
\bottomrule
\end{tabular}
\end{table}

Table \ref{nonlinearnorm} shows the second case, where the errors follow normal distributions. Our proposed method works well for approximating the best $\epsilon$-Laplacian distribution, leading to
significant improvements in the forecasting accuracy of all the simulation scenarios displayed in the Table. In particular, when the noise level is low (both $s$ and $\sigma$ are small), the superiority of the D-D approach is more prominent. For the simulation with noise settings (${\it{n}} \ 1000$, $s \ 0.7$, and $\sigma \ 0.5$), the D-D's prediction achieves an amazing improvement (MAE, $64\%$, and RMSE, $48\%$), while both the CM and $10$-CV methods each obtained only a slight increase.

\begin{table}[htbp]
\footnotesize
\centering
\caption{Nonlinear case (uniform distribution): Relative performance of the CM, $10$-CV, and D-D methods in comparison to the tuning approach.}
\label{nonlinearuniform}
\begin{tabular}{rrccccccccc}
\toprule
\multicolumn{3}{l}{Noise settings}&\multicolumn{2}{c}{Parameters}&\multicolumn{2}{c}{$\mbox{CM}$}&\multicolumn{2}{c}{$\mbox{$10$-CV}$}&\multicolumn{2}{c}{$\mbox{D-D}$}\\
  \cmidrule(lrr){1-3} \cmidrule(lrr){4-5}	\cmidrule(lr){6-7} \cmidrule(lr){8-9}\cmidrule(lr){10-11}
$\it{n}$&$s$& b &  $\hat{s}$&$\hat{\epsilon}$& $\mbox{ratio}_{\mbox{MAE}}$& $\mbox{ratio}_{\mbox{RMSE}}$& $\mbox{ratio}_{\mbox{MAE}}$& $\mbox{ratio}_{\mbox{RMSE}}$& $\mbox{ratio}_{\mbox{MAE}}$& $\mbox{ratio}_{\mbox{RMSE}}$\\
\midrule
200&	3.0&	0.8&	0.86&	0.35&	0.99&	0.99&	1.01&	1.01&	1.26&	1.22\\
400&	3.0&	0.8&	0.69&	2.15&	1.00&	1.01&	1.01&	1.01&	1.74&	1.62\\
1000&	3.0&	0.8&	0.41&	4.96&	1.00&	1.00&	1.02&	1.02&	2.16&	1.94\\
200&	3.0&	1.0&	1.08&	0.37&	1.02&	1.02&	1.01&	1.01&	1.17&	1.15\\
400&	3.0&	1.0&	0.83&	2.28&	1.01&	1.02&	1.01&	1.01&	1.66&	1.59\\
1000&	3.0&	1.0&	0.51&	4.91&	1.01&	1.01&	1.02&	1.02&	2.23&	2.06\\
200&	3.0&	1.2&	1.23&	0.80&	1.05&	1.06&	1.02&	1.02&	1.27&	1.26\\
400&	3.0&	1.2&	0.99&	2.36&	1.01&	1.02&	1.01&	1.02&	1.67&	1.59\\
1000&	3.0&	1.2&	0.62&	4.85&	1.01&	1.01&	1.02&	1.02&	2.10&	1.98\\
\hline
200&	4.0&	0.8&	1.10&	0.72&	1.02&	1.02&	1.01&	1.01&	1.24&	1.21\\
400&	4.0&	0.8&	0.87&	2.47&	1.02&	1.03&	1.02&	1.02&	1.67&	1.60\\
1000&	4.0&	0.8&	0.55&	4.87&	1.01&	1.01&	1.02&	1.02&	2.17&	2.02\\
200&	4.0&	1.0&	1.34&	0.87&	1.07&	1.07&	1.02&	1.02&	1.22&	1.20\\
400&	4.0&	1.0&	1.11&	2.25&	1.03&	1.03&	1.01&	1.02&	1.62&	1.56\\
1000&	4.0&	1.0&	0.69&	4.85&	1.01&	1.01&	1.01&	1.02&	2.07&	1.97\\
200&	4.0&	1.2&	1.66&	0.76&	1.08&	1.09&	1.02&	1.01&	1.14&	1.13\\
400&	4.0&	1.2&	1.25&	2.67&	1.04&	1.04&	1.02&	1.02&	1.54&	1.51\\
1000&	4.0&	1.2&	0.84&	4.78&	1.01&	1.02&	1.02&	1.02&	2.01&	1.93\\
\hline
200&	5.0&	0.8&	1.38&	0.63&	1.05&	1.05&	1.02&	1.01&	1.17&	1.14\\
400&	5.0&	0.8&	1.05&	2.52&	1.03&	1.04&	1.02&	1.02&	1.59&	1.53\\
1000&	5.0&	0.8&	0.66&	5.11&	1.01&	1.01&	1.02&	1.02&	2.07&	1.97\\
200&	5.0&	1.0&	1.68&	0.96&	1.07&	1.08&	1.02&	1.03&	1.15&	1.15\\
400&	5.0&	1.0&	1.33&	2.56&	1.03&	1.04&	1.01&	1.01&	1.50&	1.47\\
1000&	5.0&	1.0&	0.85&	4.97&	1.02&	1.02&	1.02&	1.02&	2.12&	2.03\\
200&	5.0&	1.2&	1.94&	1.11&	1.11&	1.10&	1.03&	1.03&	1.16&	1.14\\
400&	5.0&	1.2&	1.47&	2.86&	1.04&	1.04&	1.01&	1.01&	1.48&	1.45\\
1000&	5.0&	1.2&	0.96&	5.41&	1.03&	1.03&	1.03&	1.03&	2.00&	1.93\\
\bottomrule
\end{tabular}
\end{table}

The third nonlinear case also shows that our D-D method is an effective approach to data modeling with noises from the uniform distribution, and the simulation results are given in Table \ref{nonlinearuniform}. Obviously, two ratios from the proposed D-D method are notably greater than $1$. For instance, compared with the CM and $10$-CV methods, both ratios of the simulation from the D-D method with noise setting ${\it{n}} \ 1000$, $s \ 5.0$ and $b \ 1.2$, are nearly $200\%$ (MAE) and $193\%$ (RMSE), respectively, so our D-D method obtained a nearly twofold improvement.

From the above three types of nonlinear simulations, it can be concluded that our proposed D-D method for $\epsilon$-SVR noticeably improves the forecasting performance in nonlinear applications.

\subsection{Linear regression}

Now we consider the most popular linear model generated by the following:
\begin{equation}
y_i=\beta_0+\beta_1 \cdot x_i+ s u_i, \quad i=1, 2, \ldots,n,
\end{equation}
where $\beta_{0}=1$ and $x_i$ is generated from the normal distribution $N(0, 1)$. Considering different noise levels for all simulations, we set $\beta_1$ as $2$, $2$, and $1$ for noises generated from the $\epsilon$-Laplacian distribution, normal distribution, and uniform distribution, respectively. In addition, the kernel of the $\epsilon$-SVR is the linear function. All simulations are implemented $100$ times to record the average performance. The linear simulation results for the $\epsilon$-Laplacian distribution, normal distribution $N(0,\sigma^2)$, and uniform distribution $unif[-b, b]$ are listed in Table \ref{linearLaplacian}, Table \ref{linearnorm} and Table \ref{linearuniform}, respectively.

\begin{table}[htbp]
\footnotesize
\centering
\caption{Linear case ($\epsilon$-Laplacian distribution): Relative performance of the CM, $10$-CV, and D-D methods in comparison to the tuning approach.}
\label{linearLaplacian}
\begin{tabular}{cccccccccccc}
\toprule
\multicolumn{4}{l}{Noise settings}&\multicolumn{2}{c}{Parameters}&\multicolumn{2}{c}{$\mbox{CM}$}&\multicolumn{2}{c}{$\mbox{$10$-CV}$}&\multicolumn{2}{c}{$\mbox{D-D}$}\\
\cmidrule(lrrr){1-4} \cmidrule(lrr){5-6} \cmidrule(lr){7-8} \cmidrule(lr){9-10}\cmidrule(lr){11-12}
$\it{n}$&$s$& $\epsilon$&$R^2$& $\hat{s}$&$\hat{\epsilon}$& $\mbox{ratio}_{\mbox{MAE}}$& $\mbox{ratio}_{\mbox{RMSE}}$& $\mbox{ratio}_{\mbox{MAE}}$& $\mbox{ratio}_{\mbox{RMSE}}$& $\mbox{ratio}_{\mbox{MAE}}$& $\mbox{ratio}_{\mbox{RMSE}}$ \\
\midrule
100&	0.5&	0.8&	0.87&	0.51&	0.71&	0.80&	0.79&	0.99&	0.98&	1.13&	1.13\\
200&	0.5&	0.8&	0.87&	0.51&	0.71&	0.81&	0.83&	0.97&	0.97&	1.21&	1.21\\
300&	0.5&	0.8&	0.87&	0.49&	0.86&	1.39&	1.35&	1.14&	1.12&	1.46&	1.41\\
100&	0.5&	1.0&	0.85&	0.52&	0.74&	0.96&	0.96&	1.07&	1.06&	1.07&	1.09\\
200&	0.5&	1.0&	0.85&	0.51&	0.93&	1.18&	1.16&	1.16&	1.15&	1.26&	1.25\\
300&	0.5&	1.0&	0.86&	0.50&	0.95&	1.15&	1.14&	1.08&	1.08&	1.23&	1.22\\
100&	0.5&	1.2&	0.86&	0.52&	1.12&	0.99&	1.03&	1.08&	1.11&	1.30&	1.32\\
200&	0.5&	1.2&	0.86&	0.50&	1.21&	1.34&	1.35&	1.16&	1.15&	1.38&	1.38\\
300&	0.5&	1.2&	0.84&	0.52&	1.07&	1.20&	1.19&	1.08&	1.07&	1.33&	1.31\\
\hline
100&	1.0&	0.8&	0.65&	0.97&	0.76&	0.98&	0.96&	1.08&	1.07&	1.33&	1.28\\
200&	1.0&	0.8&	0.62&	1.01&	0.73&	0.96&	0.92&	1.04&	0.99&	1.52&	1.53\\
300&	1.0&	0.8&	0.62&	1.00&	0.76&	1.19&	1.20&	1.10&	1.11&	1.20&	1.21\\
100&	1.0&	1.0&	0.61&	1.00&	0.98&	0.76&	0.75&	1.02&	1.03&	1.20&	1.18\\
200&	1.0&	1.0&	0.61&	1.02&	0.95&	1.18&	1.16&	1.11&	1.10&	1.31&	1.28\\
300&	1.0&	1.0&	0.60&	1.01&	0.95&	1.30&	1.27&	1.15&	1.15&	1.52&	1.50\\
100&	1.0&	1.2&	0.58&	0.99&	1.27&	1.32&	1.27&	1.12&	1.11&	1.39&	1.37\\
200&	1.0&	1.2&	0.57&	1.00&	1.17&	1.46&	1.42&	1.22&	1.19&	1.55&	1.50\\
300&	1.0&	1.2&	0.58&	1.02&	1.14&	1.37&	1.36&	1.12&	1.10&	1.62&	1.60\\
\hline
100&	1.5&	0.8&	0.43&	1.48&	0.78&	0.89&	0.86&	1.09&	1.07&	1.28&	1.27\\
200&	1.5&	0.8&	0.42&	1.47&	0.77&	1.04&	1.05&	1.03&	1.03&	1.27&	1.28\\
300&	1.5&	0.8&	0.42&	1.49&	0.79&	1.19&	1.20&	1.21&	1.22&	1.25&	1.27\\
100&	1.5&	1.0&	0.42&	1.44&	1.09&	1.07&	1.06&	1.09&	1.09&	1.24&	1.23\\
200&	1.5&	1.0&	0.41&	1.49&	0.99&	1.23&	1.23&	1.14&	1.13&	1.37&	1.35\\
300&	1.5&	1.0&	0.40&	1.48&	1.05&	1.25&	1.23&	1.14&	1.13&	1.30&	1.26\\
100&	1.5&	1.2&	0.38&	1.53&	1.15&	1.37&	1.38&	1.30&	1.30&	1.73&	1.74\\
200&	1.5&	1.2&	0.38&	1.60&	1.04&	1.17&	1.20&	1.07&	1.05&	1.44&	1.40\\
300&	1.5&	1.2&	0.38&	1.50&	1.19&	1.32&	1.29&	1.11&	1.11&	1.68&	1.69\\
\bottomrule
\end{tabular}
\end{table}

First, in the linear simulation for residuals generated from the $\epsilon$-Laplacian distribution, the estimated insensitive parameter $\hat{\epsilon}$ and the estimated scale parameter  $\hat{s}$ all approximate to the real settings with our D-D method in different noise levels, as shown in Table \ref{linearLaplacian}. For comparison of the accuracy for the forecasting performance, in the linear regression with ${{\it{n}}=300}$ and $R^2=0.38$, our proposed D-D SVR performed better than the CM and the $10$-CV, with a more than $68\%$ improvement with MAE  and a $69\%$ improvement with RMSE. Overall, our D-D method can precisely improve forecasting performance by auto-adapting the insensitive parameter.

\begin{table}[htbp]
\footnotesize
\centering
\caption{Linear case (normal distribution): Relative performance of the CM, $10$-CV, and D-D methods in comparison to the tuning approach.}
\label{linearnorm}
\begin{tabular}{cccccccccccc}
\toprule
\multicolumn{4}{l}{Noise settings}&\multicolumn{2}{c}{Parameters}&\multicolumn{2}{c}{$\mbox{CM}$}&\multicolumn{2}{c}{$\mbox{$10$-CV}$}&\multicolumn{2}{c}{$\mbox{D-D}$}\\
\cmidrule(lrrr){1-4} \cmidrule(lrr){5-6} \cmidrule(lr){7-8} \cmidrule(lr){9-10}\cmidrule(lr){11-12}
$\it{n}$&$s$& $\sigma$&$R^2$&  $\hat{s}$&$\hat{\epsilon}$ &$\mbox{ratio}_{\mbox{MAE}}$& $\mbox{ratio}_{\mbox{RMSE}}$& $\mbox{ratio}_{\mbox{MAE}}$& $\mbox{ratio}_{\mbox{RMSE}}$& $\mbox{ratio}_{\mbox{MAE}}$& $\mbox{ratio}_{\mbox{RMSE}}$\\
\midrule
100&	1.0&	0.8&	0.86&	0.48&	1.28&	1.13&	1.13&	1.08&	1.09&	1.27&	1.27\\
200&	1.0&	0.8&	0.88&	0.46&	1.39&	1.09&	1.08&	1.05&	1.04&	1.37&	1.34\\
300&	1.0&	0.8&	0.86&	0.47&	1.41&	1.01&	1.04&	1.03&	1.06&	1.26&	1.27\\
100&	1.0&	1.0&	0.81&	0.59&	1.20&	0.95&	0.95&	1.06&	1.08&	1.17&	1.18\\
200&	1.0&	1.0&	0.79&	0.58&	1.43&	1.17&	1.20&	1.11&	1.13&	1.35&	1.36\\
300&	1.0&	1.0&	0.80&	0.58&	1.39&	1.10&	1.10&	0.99&	1.00&	1.34&	1.35\\
100&	1.0&	1.2&	0.74&	0.72&	1.25&	0.85&	0.81&	0.97&	0.97&	1.35&	1.32\\
200&	1.0&	1.2&	0.74&	0.73&	1.26&	1.03&	1.04&	1.00&	1.00&	1.12&	1.12\\
300&	1.0&	1.2&	0.75&	0.68&	1.51&	1.09&	1.09&	1.01&	1.00&	1.21&	1.21\\
\hline
100&	1.5&	0.8&	0.75&	0.71&	1.35&	1.04&	1.03&	1.25&	1.24&	1.22&	1.20\\
200&	1.5&	0.8&	0.73&	0.71&	1.32&	1.03&	1.03&	1.00&	1.00&	1.22&	1.19\\
300&	1.5&	0.8&	0.73&	0.70&	1.38&	1.09&	1.09&	1.03&	1.02&	1.27&	1.28\\
100&	1.5&	1.0&	0.67&	0.85&	1.49&	0.79&	0.79&	1.10&	1.10&	1.65&	1.68\\
200&	1.5&	1.0&	0.64&	0.85&	1.47&	1.22&	1.19&	1.09&	1.08&	1.34&	1.33\\
300&	1.5&	1.0&	0.64&	0.86&	1.48&	1.17&	1.19&	1.17&	1.19&	1.37&	1.39\\
100&	1.5&	1.2&	0.56&	1.03&	1.57&	1.09&	1.10&	1.05&	1.04&	1.20&	1.21\\
200&	1.5&	1.2&	0.55&	1.03&	1.48&	1.04&	1.05&	0.97&	0.96&	1.26&	1.25\\
300&	1.5&	1.2&	0.56&	1.03&	1.41&	1.16&	1.16&	1.09&	1.08&	1.24&	1.25\\
\hline
100&	2.0&	0.8&	0.62&	0.90&	1.48&	1.23&	1.20&	0.98&	0.97&	1.25&	1.22\\
200&	2.0&	0.8&	0.61&	0.88&	1.65&	1.15&	1.17&	1.05&	1.04&	1.46&	1.46\\
300&	2.0&	0.8&	0.61&	0.91&	1.54&	1.14&	1.12&	1.14&	1.11&	1.43&	1.39\\
100&	2.0&	1.0&	0.52&	1.13&	1.50&	1.07&	1.09&	1.03&	1.04&	1.21&	1.23\\
200&	2.0&	1.0&	0.51&	1.15&	1.42&	1.12&	1.12&	1.03&	1.03&	1.35&	1.32\\
300&	2.0&	1.0&	0.51&	1.11&	1.54&	1.14&	1.12&	1.08&	1.07&	1.45&	1.41\\
100&	2.0&	1.2&	0.43&	1.37&	1.43&	1.51&	1.62&	1.14&	1.15&	1.82&	1.94\\
200&	2.0&	1.2&	0.42&	1.35&	1.48&	1.20&	1.19&	1.05&	1.06&	1.22&	1.22\\
300&	2.0&	1.2&	0.40&	1.36&	1.52&	0.98&	1.00&	1.02&	1.02&	1.01&	1.03\\
\bottomrule
\end{tabular}
\end{table}

The second linear simulation, shown Table \ref{linearnorm}, is the regression with noises from the normal distribution $N(0, \sigma^2)$. The simulation results show that with $R^2$ from $0.40$ to $0.86$, all the $\mbox{ratio}_{\mbox{MAE}}$ and $\mbox{ratio}_{\mbox{RMSE}}$ for D-D are all significantly greater than 1. In other words, our proposed method can auto-recognize a limited scale and obtain a limiting insensitive parameter to approach real noises; as a result, the forecasting performance is superior. It is interesting that corresponding to the type of noise, the scale is also auto-adapted to match the most approximate $\epsilon$ in the insensitive Laplacian distribution. Therefore, our method can make $\epsilon$-SVR more efficient in the linear model with Gaussian noises.

\begin{table}[htbp]
\footnotesize
\centering
\caption{Linear case (uniform distribution): Relative performance of the CM, $10$-CV, and D-D methods in comparison to the tuning approach.}
\label{linearuniform}
\begin{tabular}{ccccccccccccc}
\toprule
\multicolumn{4}{l}{Noise settings}&\multicolumn{2}{c}{Parameters}&\multicolumn{2}{c}{$\mbox{CM}$}&\multicolumn{2}{c}{$\mbox{$10$-CV}$}&\multicolumn{2}{c}{$\mbox{D-D}$}\\
\cmidrule(lrrr){1-4} \cmidrule(lrr){5-6} \cmidrule(lr){7-8} \cmidrule(lr){9-10}\cmidrule(lr){11-12}
$\it{n}$&$s$& $b$ &$R^2$&  $\hat{s}$&$\hat{\epsilon}$ &$\mbox{ratio}_{\mbox{MAE}}$& $\mbox{ratio}_{\mbox{RMSE}}$& $\mbox{ratio}_{\mbox{MAE}}$& $\mbox{ratio}_{\mbox{RMSE}}$& $\mbox{ratio}_{\mbox{MAE}}$& $\mbox{ratio}_{\mbox{RMSE}}$\\
\midrule
100&	1.0&	0.8&	0.81&	0.19&	4.06&	1.25&	1.23&	1.11&	1.09&	2.72&	2.69\\
200&	1.0&	0.8&	0.82&	0.12&	6.29&	1.21&	1.20&	1.14&	1.14&	3.25&	3.23\\
300&	1.0&	0.8&	0.83&	0.09&	8.25&	1.00&	1.00&	1.13&	1.12&	3.14&	3.09\\
100&	1.0&	1.0&	0.77&	0.22&	4.19&	1.42&	1.43&	1.08&	1.09&	2.42&	2.46\\
200&	1.0&	1.0&	0.76&	0.14&	7.15&	1.20&	1.19&	1.13&	1.13&	3.00&	2.96\\
300&	1.0&	1.0&	0.76&	0.12&	8.15&	1.16&	1.17&	1.16&	1.17&	3.89&	3.89\\
100&	1.0&	1.2&	0.68&	0.25&	4.54&	1.47&	1.47&	1.15&	1.15&	2.56&	2.57\\
200&	1.0&	1.2&	0.67&	0.18&	6.61&	1.14&	1.16&	1.09&	1.09&	3.63&	3.65\\
300&	1.0&	1.2&	0.68&	0.14&	8.37&	1.18&	1.20&	1.19&	1.21&	3.61&	3.58\\
\hline
100&	1.5&	0.8&	0.66&	0.28&	4.19&	1.13&	1.13&	1.04&	1.03&	2.34&	2.29\\
200&	1.5&	0.8&	0.68&	0.16&	7.52&	1.07&	1.07&	1.01&	1.02&	3.02&	3.05\\
300&	1.5&	0.8&	0.68&	0.15&	8.06&	1.20&	1.20&	1.16&	1.15&	3.36&	3.32\\
100&	1.5&	1.0&	0.58&	0.33&	4.47&	1.64&	1.63&	1.09&	1.09&	2.45&	2.44\\
200&	1.5&	1.0&	0.57&	0.21&	7.30&	1.21&	1.23&	1.10&	1.10&	3.91&	3.92\\
300&	1.5&	1.0&	0.57&	0.17&	8.69&	1.09&	1.08&	1.10&	1.10&	4.08&	4.16\\
100&	1.5&	1.2&	0.47&	0.35&	5.50&	1.23&	1.20&	1.06&	1.05&	2.08&	2.07\\
200&	1.5&	1.2&	0.50&	0.24&	7.55&	1.17&	1.19&	1.05&	1.05&	3.73&	3.90\\
300&	1.5&	1.2&	0.48&	0.21&	8.53&	1.17&	1.19&	1.13&	1.13&	4.13&	4.14\\
\hline
100&	2.0&	0.8&	0.53&	0.30&	5.29&	1.42&	1.40&	1.16&	1.12&	3.06&	3.06\\
200&	2.0&	0.8&	0.54&	0.21&	7.85&	1.18&	1.18&	1.15&	1.16&	3.17&	3.20\\
300&	2.0&	0.8&	0.53&	0.19&	8.45&	1.05&	1.02&	1.06&	1.04&	4.27&	4.13\\
100&	2.0&	1.0&	0.43&	0.34&	5.88&	1.16&	1.20&	0.96&	0.98&	2.46&	2.48\\
200&	2.0&	1.0&	0.44&	0.27&	7.43&	1.05&	1.05&	1.00&	0.99&	2.85&	2.83\\
300&	2.0&	1.0&	0.43&	0.22&	9.25&	1.14&	1.14&	1.17&	1.17&	4.17&	4.18\\
100&	2.0&	1.2&	0.35&	0.46&	5.03&	1.38&	1.40&	1.08&	1.08&	3.22&	3.39\\
200&	2.0&	1.2&	0.35&	0.33&	7.58&	1.12&	1.10&	1.04&	1.02&	3.20&	3.14\\
300&	2.0&	1.2&	0.34&	0.26&	9.20&	1.09&	1.07&	1.10&	1.10&	4.32&	4.20\\
\bottomrule
\end{tabular}
\end{table}

The final simulation, shown in Table \ref{linearuniform}, illustrates that our D-D method can obtain surprisingly good improvements. This is because the ratios from our D-D method are quite large, indicating that our proposed method can model the linear model with perfect accuracy. The most interesting finding in the parameter estimation analysis is that with an increasing number of samples, our D-D method approaches approximating the $\epsilon$-Laplacian loss function by increasing $\epsilon$ and decreasing $s$; two parameter estimations will converge to limiting values. To sum up, for the noise from uniform distribution, our method is still a powerful tool for improving the linear regression forecasting.

Furthermore, for the mechanism exploration of our D-D method, compared with the CM  in linear simulations, which is motivated by the noise following the normal distribution, our D-D's forecasting performance is close, but still is better when addressing the noise from the normal distribution shown in Table \ref{linearnorm}, while in Table \ref{linearLaplacian} and Table \ref{linearuniform}, our D-D method's performance can significantly improve the forecasting accuracy. This illustrates that our D-D method can auto-adapt the parameters to approximate any unknown noise distribution and improve the SVR's performance, while the CM method focuses on the normal distribution. Moreover, the computational cost of the $10$-CV method with five alternative parameter settings is over 10 times more than our D-D method. In addition, because of the parameter setting for the cross validation, the $10$-CV method cannot guarantee its superior performance with high computational costs. Therefore, we can conclude that our D-D method can auto-adapt the $\epsilon$-Laplacian loss function to guarantee the steadiness of a linear model with high levels of accuracy. Furthermore, because it is determined by the type of noise, the scale and the insensitive parameter will converge to true values (the noise is generated from the $\epsilon$-Laplacian distribution) or limiting values (the noise is from any other distribution).

\section{Case studies}\label{case}

In the section, our D-D $\epsilon$-SVR is evaluated with five case studies: energy efficiency ($768$ samples, eight attributes, and two responses \citep{tsanas2012accurate}, yacht hydrodynamics ($308$ samples, six attributes, and one response) \citep{ortigosa2007neural}, airfoil self-noise ($1503$ samples, five attributes, and one response) \citep{lau2006neural}, concrete compressive strength ($1030$ samples, eight attributes, and one response) \citep{yeh2006analysis} from the UCI Machine Learning Repository \citep{Dua:2019}, and Boston housing prices ($506$ samples, $14$ attributes, and one response) from the StatLib collection \citep{website2}.

Each benchmark data set was randomly divided into two groups: the training set ($70\%$ of each data set) and the test set (the remaining data from each set). Then, each experiment was repeated $100$ times to obtain the average performance of our proposed SVR. Because the scale of each attribute is different, the standard normalization was applied for attribute pre-processing before the training. The general radial basic function is selected as the kernel. In addition, the $10$-CV was applied in the insensitive parameter selection with the same alternative parameter settings as the former simulations.

\begin{figure}[htpb]
\centering
\includegraphics[height=12cm,width=13cm]{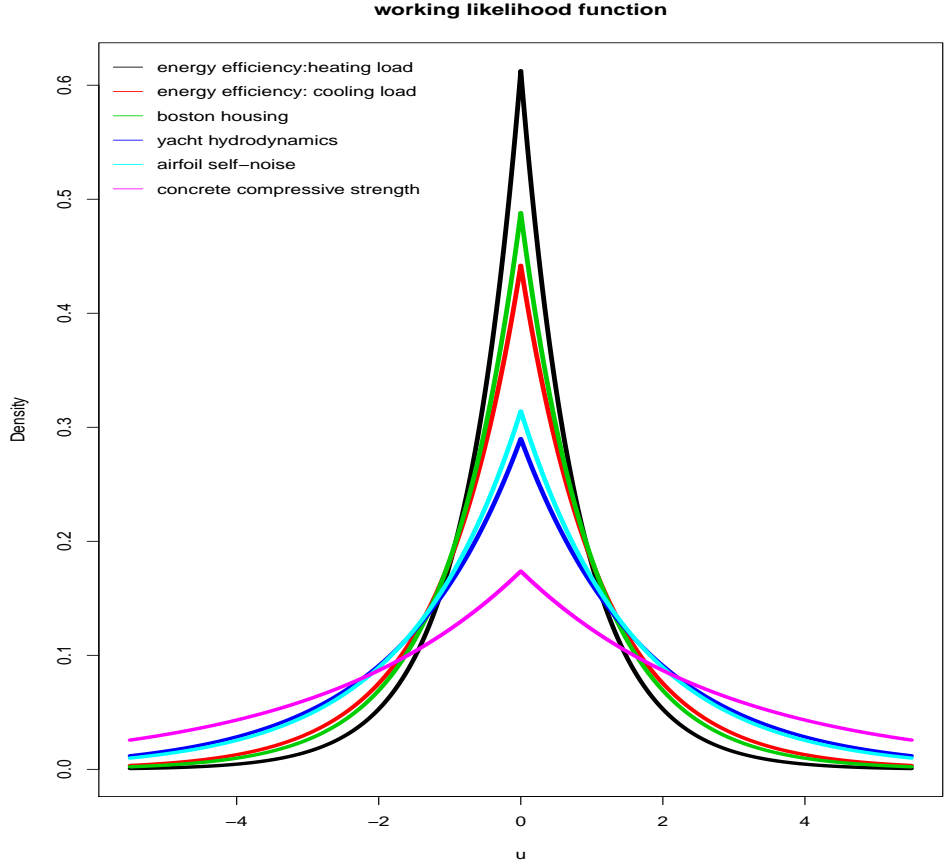}
\caption{Six working likelihood D-D functions for five case studies.}
\label{casestudy}
\end{figure}

The $\epsilon$ and $\sigma$ for the five benchmark data sets were estimated using our proposed method, and the work likelihood functions are displayed in Figure \ref{casestudy}. It is obvious that the specific $\epsilon$-Laplacian loss function was data-driven by the real data sets. Different from the original $\epsilon$-SVR, our proposed ``scale" $\epsilon$-SVR can auto-recognize the scale of noise in real data sets and self-adapt the insensitive parameter accordingly. For example, as illustrated in Figure \ref{casestudy}, the working likelihood functions for energy efficiency (heating load) and concrete compressive strength were significantly different; the scale parameter estimation for the former data set was $0.82$, while the estimation for the latter was $2.88$, as shown in Table \ref{case_study}.

\begin{table}[htbp]
\footnotesize
\centering
\caption{Results for four case studies: Relative performance of the tuning, CM, $10$-CV, and D-D methods.}
\label{case_study}
\begin{spacing}{1.5}
\begin{tabular}{llcccccccccc}
\toprule
            &\multicolumn{2}{c}{Parameters}&\multicolumn{4}{c}{MAE}&\multicolumn{4}{c}{RMSE}\\
  \cmidrule(lrr){2-3} \cmidrule(lrr){4-7}	\cmidrule(lrr){8-11}
	 &$\hat{s}$& $\hat{\epsilon}$&	tuning& CM& $\mbox{$10$-CV}$& $\mbox{D-D}$  & 	tuning&  CM& $\mbox{$10$-CV}$&$\mbox{D-D}$  \\
\midrule
\multicolumn{4}{l}{ Energy efficiency}&&&&&&&&\\
Heating Load	  &		0.82&	0.00&	1.49&	1.44&	1.49& \bf{1.08}&     2.27&	2.30&	2.27&   \bf{1.87}\\
Cooling Load	  &		1.13&	0.00&	1.84&	1.82&	1.81& \bf{1.48}&	    2.68&	2.69&	2.69&  	\bf{2.32}\\
\cline{2-11}
&&&&&&&&&&&\\
\multicolumn{4}{l}{Boston Housing}&&&&&&&&\\
                  &		1.02&	0.00&	2.38&	2.39&	2.39& \bf{2.22}&	    3.97&	3.96&	3.97&	\bf{3.53}\\
\cline{2-11}
&&&&&&&&&&&\\
\multicolumn{4}{l}{Yacht Hydrodynamics}&&&&&&&&\\
                  &		1.72&	0.00&	3.83&	4.09&	4.17& \bf{2.67}&	    6.81&	6.71&	6.70&	\bf{4.95}\\
\cline{2-11}
&&&&&&&&&&&\\
\multicolumn{4}{l}{Airfoil Self-Noise }&&&&&&&&\\
                  &		1.59&	0.00&	2.41&	2.42&	2.42& \bf{1.96}&	     3.31&	3.31&	3.31&	\bf{2.79}\\
\cline{2-11}
&&&&&&&&&&&\\
\multicolumn{4}{l}{Concrete Compressive Strength }&&&&&&&&\\
                  &		2.88&	0.00&	5.00&	5.00&	4.97& \bf{4.29}&	     6.84&	6.84&	6.85&	\bf{6.13}\\
\bottomrule
\end{tabular}
\end{spacing}
\end{table}

The prediction performance for all five cases are listed in Table \ref{case_study}. Obviously, our proposed method can dramatically improve the accuracy of predictions based on the ratios. The most obvious cases are the MAE (tuning $3.83$ vs. CM $4.09$ vs. $10$-CV $4.17$ vs. D-D $2.67$) and RMSE (tuning $6.81$ vs. CM $6.71$ vs. $10$-CV $6.70$ vs. D-D $4.95$) for the yacht hydrodynamics. Compared with the tuning, $10$-CV, and CM methods, the MAE and RMSE in the rest of the data sets (energy efficiency, Boston housing, airfoil self-noise, and concrete compressive strength) achieved around $10\%$ improvements.

To summarize, our proposed D-D method can auto-adapt the insensitive parameter in the $\epsilon$-Laplacian distribution approach to the real noise distribution; this means our working likelihood method can push the $\epsilon$-Laplacian density function to seek the approximate likelihood function. As a result, our D-D SVR has an excellent performance in real applications.

\section{Conclusion}\label{conclusion}

The SVR with $\epsilon$-Laplacian loss distribution is a mainstream algorithm for regression modeling, where the insensitive parameter $\epsilon$ determines the support vector. However, to date, after inputs and target scaling, three types of strategies for parameter selection are used: the $\it{k}$-cross validation, which requires huge computational costs, the tuning parameter, which cannot make the SVR work more efficiently, and the empirical statistical estimation, the CM method that is based on normal distribution with some empirical settings. Obviously, the mentioned parameter settings are not the most appropriate hyper-parameters for SVR in most conditions, so, in this paper, we propose optimization of the insensitive parameter based on the working likelihood function developed by \citet{wang2007robust}, which is a D-D method, to estimate appropriate hyper-parameters for finding the most appropriate $\epsilon$-Laplacian distribution to the real noise distribution in order to guarantee generalization in test sets. In addition, the D-D vector regression is standardized by the scale of the noise in a more meaningful field. In nonlinear and linear simulations conducted with different types of noises ($\epsilon$-Laplacian distribution, normal distribution, and uniform distribution), our proposed method demonstrated that it can automatically estimate the scale and the insensitive parameter. As a result, our D-D SVR showed significantly improved forecasting accuracy in the test sets. Moreover, our D-D algorithm can estimate the approximate likelihood function in five real benchmark applications, and furthermore, the proposed method had dramatically improved performance in unknown sets. Therefore, our proposed D-D SVR is a more intelligent and powerful technique for the regression problem. Furthermore, in machine learning modeling, our D-D method using the framework of working likelihood is a viable general strategy for parameter estimations in different loss functions.

\section*{Declaration of interest}

The authors declare no conflict interest.

\section*{Acknowledgements}

Computational (and/or data visualisation) resources and services used in this work were provided by the HPC and Research Support Group, Queensland University of Technology, Brisbane, Australia. This work was supported in part by the ARC Centre of Excellence for Mathematical and Statistical Frontiers. This work was supported by the Australian Research Council project DP160104292.

\bibliographystyle{apacite}

\bibliography{RefSVR}

\end{document}